\documentclass[runningheads,a4paper]{llncs}

\usepackage{amssymb}
\usepackage{graphicx}
\usepackage{url}

\usepackage{amsmath}
\usepackage{algorithmic}
\usepackage{algorithm}
\usepackage{subfigure}
\usepackage{array}
\usepackage{tabularx}
\usepackage{multirow}

\newcommand{\bfx}{{\textbf{x}}}
\newcommand{\bfv}{{\textbf{v}}}

\newcommand{\bfw}{{\textbf{w}}}

\newcommand{\bfxi}{{\boldsymbol{\xi}}}

\newcommand{\bfalpha}{{\boldsymbol{\alpha}}}

\newcommand{\bfepsilon}{{\boldsymbol{\epsilon}}}
\newcommand{\bfdelta}{{\boldsymbol{\delta}}}

\newcommand{\keywords}[1]{\par\addvspace\baselineskip
\noindent\keywordname\enspace\ignorespaces#1}

\begin{document}

\mainmatter  

\title{Representing data by sparse combination of contextual data points for classification}

\titlerunning{Representing data by sparse combination}

\author{Jingyan Wang$^{1,2,3}$
\and
Yihua Zhou$^4$
\and
Ming Yin$^5$
\and
Shaochang Chen$^5$
\and
Benjamin Edwards$^6$}

\authorrunning{J. Wang, et al.}

\institute{
National Time Service Center, Chinese Academy of Sciences, Xi' an, Shaanxi 710600 , China
\and
Graduate University of Chinese Academy of Sciences, Beijing 100049, China
\and
Provincial Key Laboratory for Computer Information Processing Technology, Soochow University Suzhou 215006, China\\
\email{jingbinwang1@outlook.com}
\and
Department of mechanical engineering and mechanics, Lehigh University, Bethlehem, PA 18015, USA
\and
Electronic Engineering College, Naval University of Engineering, Wuhan 430033,  China
\and
Department of Computer Science, Sam Houston State University, Huntsville, TX 77341, USA\\
\email{benjamin.edwards1@hotmail.com}
}

\maketitle

\begin{abstract}
In this paper, we study the problem of using contextual data points of a data point for its classification problem. We propose to represent a data point as the sparse linear reconstruction of its context, and learn the sparse context to gather with a linear classifier in a supervised way to increase its discriminative ability. We proposed a novel formulation for context learning, by modeling the learning of context reconstruction coefficients  and classifier in a unified objective. In this objective, the reconstruction error is minimized and the coefficient sparsity is encouraged. Moreover, the hinge loss of the classifier is minimized and the complexity of the classifier is reduced. This objective is optimized by an alternative strategy in an iterative algorithm. Experiments on three benchmark data set show its advantage over  state-of-the-art context-based data representation and classification methods.
\keywords{Pattern classification,
Context learning,
Nearest neighbors, and
Sparse regularization}
\end{abstract}

\section{Introduction}
\label{sec:intro}

Pattern classification is a major problem in machine learning research \cite{Xu20121205,Guo20121893,He2013793,Tian20141007}. The two most important topics of pattern classification are data representation and classifier learning. Zhang et al. proposed an efficient multi-model classifier for large scale Bio-sequence localization prediction \cite{zhang2009bayesian}. Zhang et al. developed and optimized association rule mining algorithms and implemented them on paralleled micro-architectural platforms \cite{zhang2013accelerating,zhang2011gpapriori}.
Most data representation and classification methods are based on single data point. When one data point is considered for representation and classification, all other data points are ignored. However, the other data points other than the data point under consideration, which are called contextual data points, may play important roles in its representation and classification.
It is necessary to explore the contexts of data points when they are represented and/or classified. In this paper, we investigate the problem of learning effective representation of a data point from its context guided by its class label, and proposed a novel supervised context learning method using sparse regularization and linear classifier learning formulation.

We propose a novel method to explore the context of a data point, and use it to represent it. We use its $k$ nearest neighbors as its context, and try to reconstruct it by the data points in its context. The reconstruction errors are imposed to be spares. Moreover, the reconstruction result is used as the new representation of this data point.
We apply a linear function to predict its class label from the sparse reconstruction of its context. The motivation of this contribution is that for each data point, only a few data points in its context is of the same class as itself. To find the critical contextual data points, we proposed to learn the classifier together with she sparse context. We mode this problem as a minimization problem. In this problem, the context reconstruction error, reconstruction sparsity, classification error, and classifier complexity are minimized simultaneously.
We also problem a novel iterative algorithm to solve this minimization problem. We first reformulate it as ist Lagrange formula, and the use an alterative optimization method to solve it.

This paper is organized as follows. In section \ref{sec:method}, we introduce the proposed method. In section \ref{sec:experiment}, we evaluate the proposed method experimentally. In section \ref{sec:conclusion}, this paper is concluded with future works.

\section{Proposed method}
\label{sec:method}

We consider a binary classification problem, and a training set of $n$ data points are given as $\{(\bfx_i,y_i)\}_{i=1}^n$, where $\bfx_i\in \mathbb{R}^d$ is a $d$-dimensional feature vector of the $i$-th data point, and $y_i\in \{+1,-1\}$ is the class label of the $i$-th point.
To learn from the context of the $i$-th data point, we find its $k$ nearest neighbors and denote
them as $\{\bfx_{ij}\}_{j=1}^k$, where $\bfx_{ij}$ is the $j$-th nearest neighbor of the $i$-th point.
They are further organized as a $d\times k$ matrix $X_i=[\bfx_{i1},\cdots,\bfx_{ik} ] \in R^{d\times {k}}$, where the $j$-th column is $\bfx_{ij}$. We represent $\bfx_i$ by linearly reconstructing it from its contextual points as

\begin{equation}
\begin{aligned}
\bfx_i \approx \widehat{\bfx}_i
= \sum_{j=1}^k \bfx_{ij} v_{ij}
= X_i \bfv_i
\end{aligned}
\end{equation}
where $\widehat{\bfx}_i$ is its reconstruction, and $v_{ij}$ is the reconstruction coefficient
of the $j$-th nearest neighbor.
$\bfv_i=[v_{i1},\cdots,v_{ik}]^\top \in \mathbb{R}^k$ is the reconstruction coefficient vector of the $i$-th data point. The reconstruction coefficient vectors of all the training points are organized in reconstruction coefficient matrix
$V=[\bfv_1,\cdots,\bfv_n] \in \mathbb{R}^{k\times n}$, with its $i$-th column as $\bfv_i$. To solve the reconstruction coefficient vectors, we propose the following minimization problem,

\begin{equation}
\label{equ:reconstruction}
\begin{aligned}
\min_{V}
~
&\left \{
\beta \sum_{i=1}^n  \|\bfx_i - X_i \bfv_i\|_2^2 +  \gamma \sum_{i=1}^n  \|\bfv_i\|_1 \right \},
\end{aligned}
\end{equation}
where $\beta$ and $\gamma$ are trade-off parameters. In the objective of this problem, the first term is to minimize the reconstruction error measured by a squared $\ell_2$ norm penalty between $\bfx_i$ and $X_i \bfv_i$, and the second term is a $\ell_1$ norm penalty to the
contextual reconstruction coefficient vector $\bfv_i$.

We design a classifier to classify the $i$-th data point,

\begin{equation}
\begin{aligned}
f(\widehat{\bfx}_i) = \bfw^\top  \widehat{\bfx}_i
=\bfw^\top X_i \bfv_i
\end{aligned}
\end{equation}
where $\bfw\in \mathbb{R}^d$ is the classifier parameter vector.
The following optimization problem is proposed to learn $\bfw$,

\begin{equation}
\label{equ:classification}
\begin{aligned}
\min_{\bfw,V,\bfxi}
~
&\left \{ \frac{1}{2}\| \bfw \|^2_2 +
\alpha \sum_{i=1}^n  \xi_i \right \}\\
s.t.~
&
1- y_i\left (\bfw^\top X_i \bfv \right ) \leq \xi_i, \xi_i \geq 0, i=1,\cdots, n,
\end{aligned}
\end{equation}
where $\frac{1}{2}\| \bfw \|^2_2 $ is the the squared $\ell_2$ norm regularization term to reduce the complexity
of the classifier, $\xi_i$ is the slack variable for the hinge loss of the $i$-th training point, $\bfxi=[\xi_1,\cdots,\xi_n]^\top$ and $\alpha$ is a tradeoff parameter.

The overall optimization problem is obtained by combining the problems in both (\ref{equ:reconstruction}) and (\ref{equ:classification}) as

\begin{equation}
\label{equ:objective}
\begin{aligned}
\min_{\bfw,V,\bfxi}
~
&\left \{ \frac{1}{2}\| \bfw \|^2_2 +
\alpha \sum_{i=1}^n  \xi_i + \beta \sum_{i=1}^n  \|\bfx_i - X_i \bfv_i\|_2^2 +  \gamma \sum_{i=1}^n  \|\bfv_i\|_1
\right \} \\
s.t.~
&
1- y_i\left (\bfw^\top X_i \bfv \right ) \leq \xi_i, \xi_i \geq 0, i=1,\cdots, n.
\end{aligned}
\end{equation}
According to the dual theory of optimization, the following dual optimization problem is obtained,

\begin{equation}
\label{equ:dual}
\begin{aligned}
\max_{\bfdelta,\bfepsilon}&\min_{\bfw,V,\bfxi}
~\left \{
\frac{1}{2}\| \bfw \|^2_2 +
\alpha \sum_{i=1}^n  \xi_i + \beta \sum_{i=1}^n  \|\bfx_i - X_i \bfv_i\|_2^2 +  \gamma \sum_{i=1}^n  \|\bfv_i\|_1
\right .\\
&
\left .
+\sum_{i=1}^n \delta_i \left(1- y_i\left (\bfw^\top X_i \bfv_i \right ) - \xi_i \right)
-\sum_{i=1}^n \epsilon_i \xi_i
\right \},\\
s.t.~
&
\bfdelta\geq 0,\bfepsilon\geq 0,
\end{aligned}
\end{equation}
where $\bfdelta=[\delta_1,\cdots,\delta_n]^\top$, and $\bfepsilon=[\epsilon_1,\cdots,\epsilon_n]^\top$ are Lagrange multipliers. By setting the partial derivative of $\mathcal{L}$ with regard to $\bfw$  and $\xi_i$ to zeros, we have

\begin{equation}
\label{equ:w}
\begin{aligned}
&\bfw = \sum_{i=1}^n \delta_i  y_i  X_i \bfv_i.\\
&\alpha-\delta_i=\epsilon_i\\
&\Rightarrow
\alpha \geq \delta_i.
\end{aligned}
\end{equation}
We substitute (\ref{equ:w}) to (\ref{equ:dual})to eliminate $\bfw$ and $\bfdelta$,

\begin{equation}
\label{equ:objective1}
\begin{aligned}
\max_{\bfdelta}&\min_{V}
~
\left \{
-\frac{1}{2} \sum_{i,j=1}^n \delta_i  \delta_j y_i y_j \bfv_i ^\top X_i^\top   X_j \bfv_j
+ \beta \sum_{i=1}^n  \|\bfx_i - X_i \bfv_i\|_2^2
\right. \\
&\left .+  \gamma \sum_{i=1}^n  \|\bfv_i\|_1
+\sum_{i=1}^n \delta_i
\right \}
\\
s.t.~
&
\bfalpha \geq \bfdelta \geq 0.
\end{aligned}
\end{equation}
where $\bfalpha=[\alpha,\cdots,\alpha]^\top$ is a $n$ dimensional vector of all $\alpha$ elements. We solve this problem with the alternate optimization strategy. In each iteration of an iterative algorithm, we fix $\bfdelta$ first to solve $V$, and then fix $V$ to solve $\bfdelta$.

\begin{description}
\item[Solving $V$]
When $\bfdelta$ is fixed and only $V$ is considered, we solve $\bfv_i|_{i=1}^n$ one by one, (\ref{equ:objective1}) is further reduced to

\begin{equation}
\label{equ:v}
\begin{aligned}
\min_{\bfv_i}
~
&\left \{
-\frac{1}{2} \sum_{i,j=1}^n \delta_i  \delta_j y_i y_j \bfv_i ^\top X_i^\top   X_j \bfv_j
+ \beta   \|\bfx_i - X_i \bfv_i\|_2^2 +  \gamma   \|\bfv_i\|_1 \right \}.
\end{aligned}
\end{equation}
This problem could be solved efficiently by the modified feature-sign search algorithm
proposed by Gao et al. \cite{gao2013laplacian}.

\item[Solving $\bfdelta$]
When $V$ is fixed and only $\bfdelta$ is considered, the problem in (\ref{equ:objective1}) is reduced to

\begin{equation}
\label{equ:delta}
\begin{aligned}
\max_{\bfdelta}
~
&
\left \{
-\frac{1}{2} \sum_{i,j=1}^n  \delta_i  \delta_j y_i y_j \bfv_i ^\top X_i^\top   X_j \bfv_j
+\sum_{i=1}^n \delta_i \right \}\\
s.t.~
&
\bfalpha \geq \bfdelta \geq 0.
\end{aligned}
\end{equation}
This problem is a typical constrained quadratic programming (QP) problem, and it can be solved efficiently
by the active set algorithm.

\end{description}

\section{Experiments}
\label{sec:experiment}

In this section, we evaluate the proposed supervised sparse context learning (SSCL) algorithm on several benchmark data sets.

\subsection{Experiment setup}

In the experiments, we used three date sets, which are introduced as follows:

\begin{itemize}
\item \textbf{MANET loss data set}:
The packet losses of the receiver in mobile Ad hoc networks (MANET) can be classified into three types, which are wireless random errors caused losses, the route change losses induced by node mobility and network congestion. We collect 381 data points for the congestion loss, 458 for the route change loss, and 516 data points for the wireless error loss for this data set. Thus in the data set, there are 1355 data points in total. To extract the feature vector each data point, we calculate 12 features from each data point as in \cite{Deng2009}, and concatenate them to form a vector.

\item \textbf{Twitter data set}: The second data set is a Twitter data set. The target of this data set is to predict the gender of the twitter user, male or female, given one of his/her Twitter massage. We collected 53,971 twitter massages in total, and among them there are 28,012 messages sent by male users, and 25,959 messages sent by female users. To extract features from each Twitter message, we extract Term features, linguistic features, and medium diversity features as gender-specific features as in  \cite{Huang2014488}.

\item \textbf{Arrhythmia data set}: The third data set is publicly available at http://arc\\
    hive.ics.uci.edu/ml/datasets/Arrhythmia. In this data set, there are 452 data points, and they belongs to 16 different classes. Each data point has a feature vector of 279 features.

\end{itemize}

To conduct the experiments, we used the 10-fold cross validation.

\subsection{Experimental Results}

\begin{figure}[!htb]
  \centering
  \subfigure[MANET loss data set]{
  \includegraphics[width=0.3\textwidth]{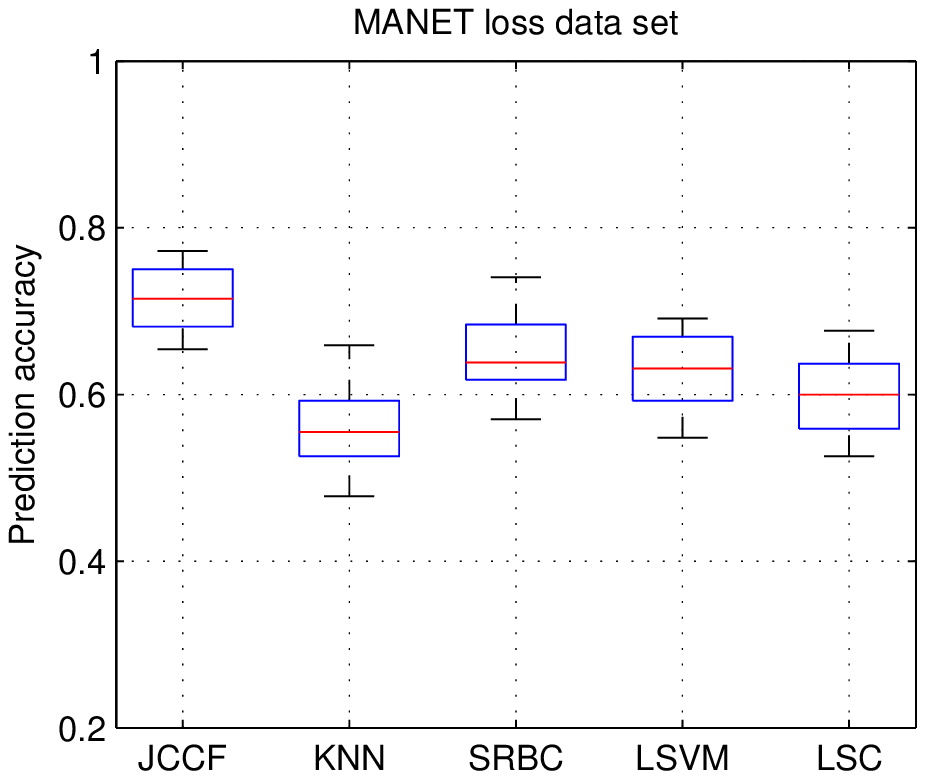}}
  \subfigure[Twitter data set]{
  \includegraphics[width=0.3\textwidth]{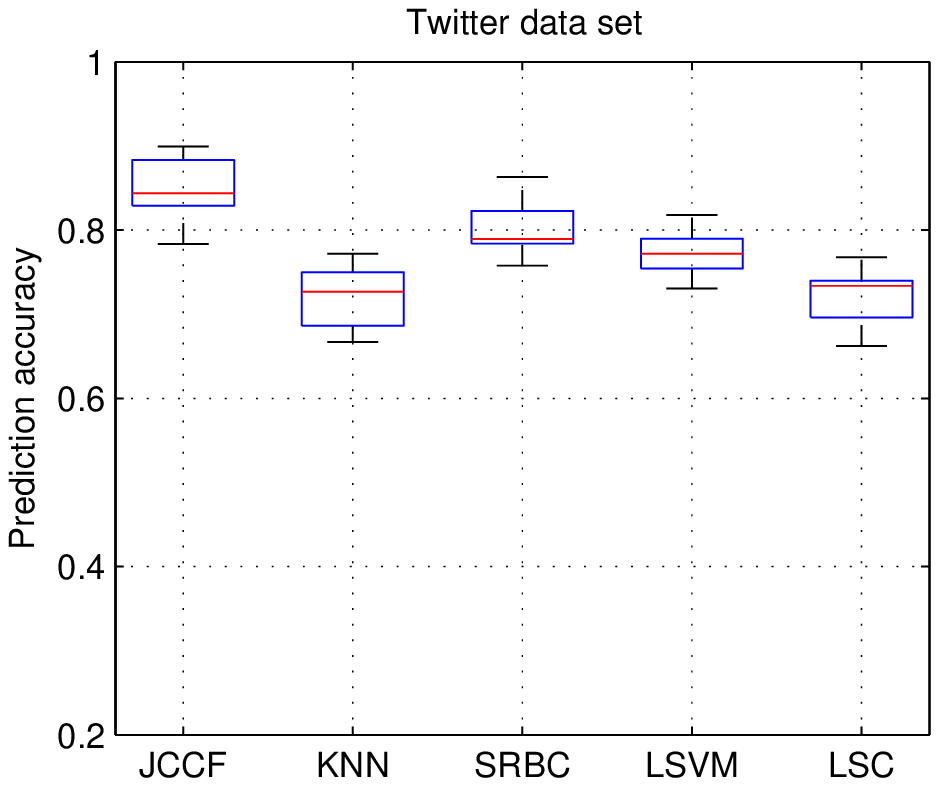}}
  \subfigure[Arrhythmia data set]{
  \includegraphics[width=0.3\textwidth]{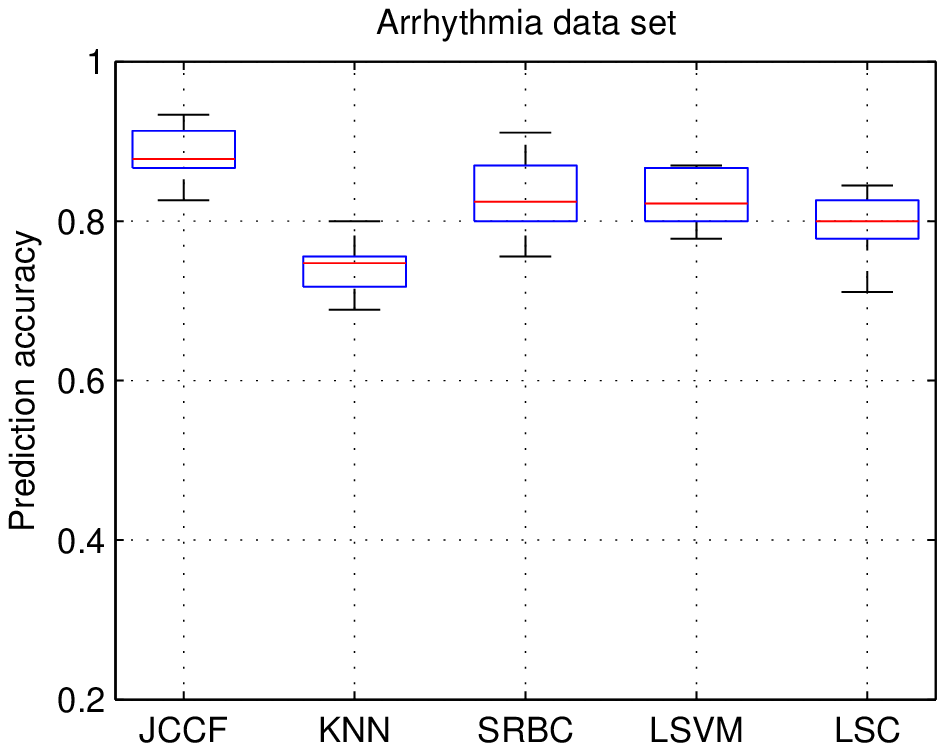}}
  \caption{Boxplots of prediction accuracy of different context-based algorithms.}
  \label{fig:compare}
\end{figure}

Since the proposed algorithm is a context-based classification and sparse representation method, we compared the proposed algorithm to three popular context-based classifiers, and one context-based sparse representation method. The three context-based classifiers are traditional $k$-nearest neighbor classifier (KNN), sparse representation based classification (SRBC) \cite{wright2009robust},and Laplacian support vector machine (LSVM) \cite{melacci2011laplacian}. The context-based sparse representation method is Gao et al.'s Laplacian sparse coding (LSC) \cite{gao2010local}. The boxplots of the 10-fold cross validation of the compared algorithms are given in figure \ref{fig:compare}. From the figures, we can see that the proposed method SSCL outperforms all the other methods on all three data sets. The second best method is SRBC, which also uses sparse context to represent the data point. This is a strong evidence that learning a supervised sparse context is critical for classification problem.

\subsubsection{Sensitivity to parameters}

\begin{figure}[!htb]
  \centering
\subfigure[$\alpha$]{
\label{fig:alpha}
\includegraphics[width=0.3\textwidth]{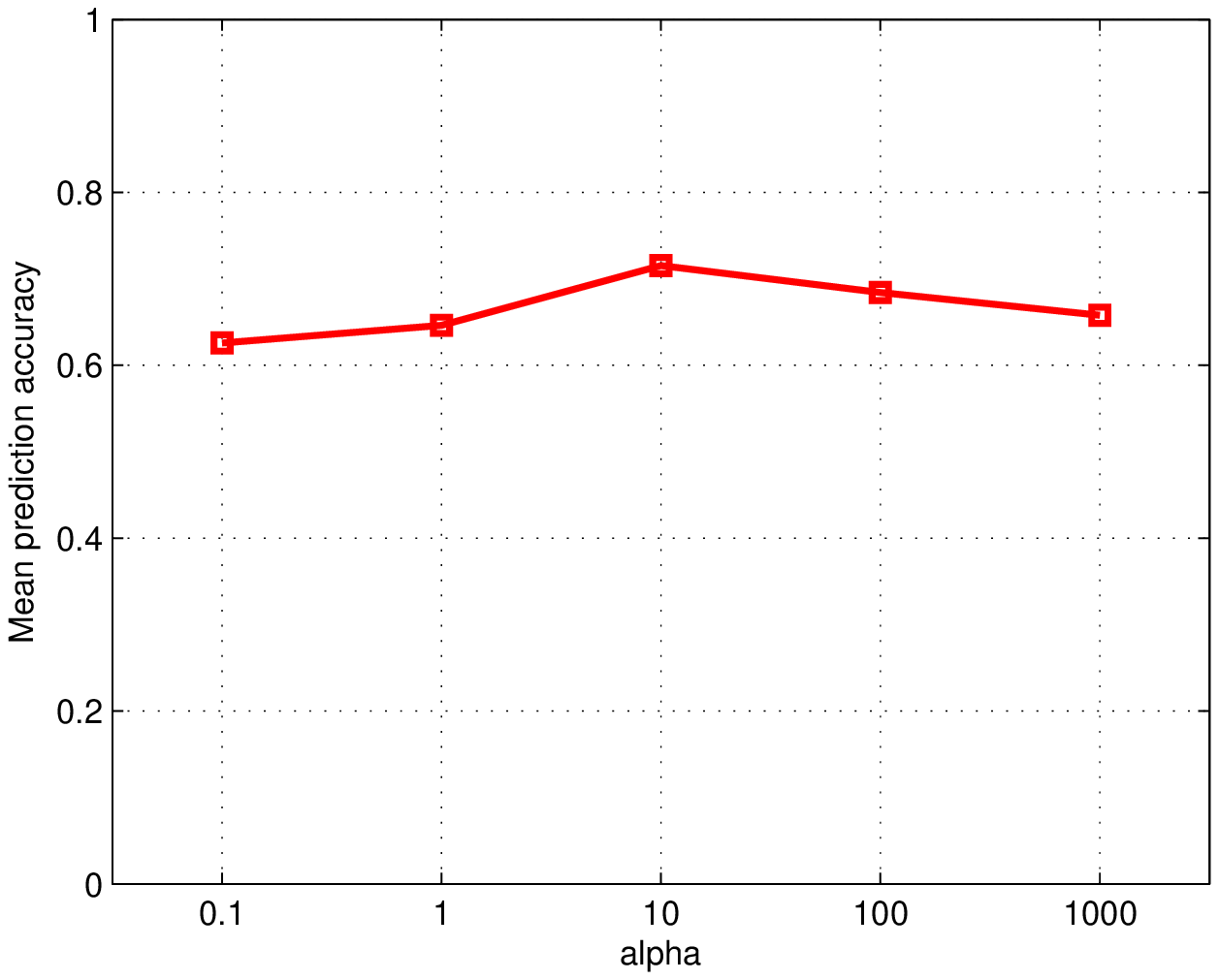}}
\subfigure[$\beta$]{
\label{fig:beta}
\includegraphics[width=0.3\textwidth]{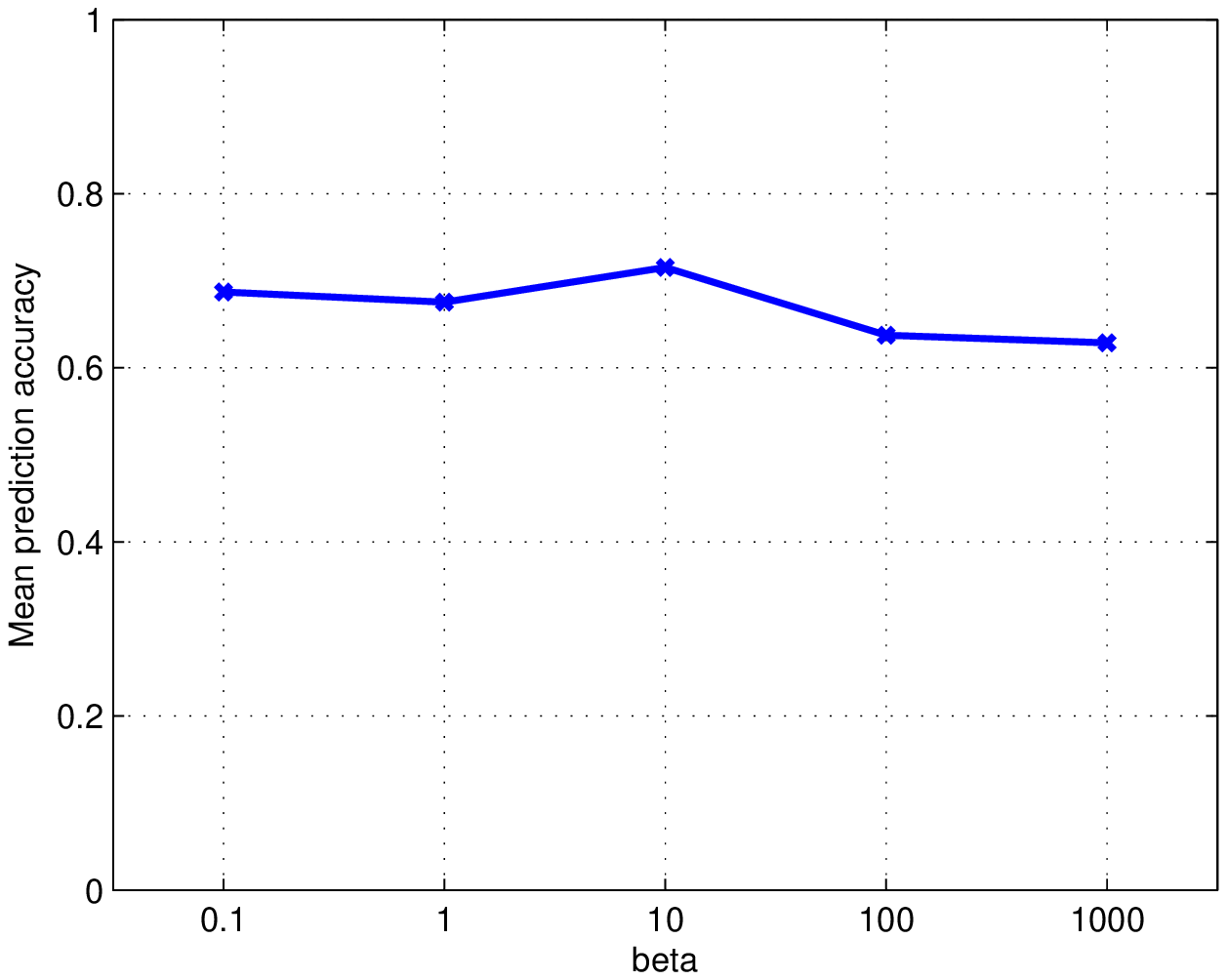}}
\subfigure[$\gamma$]{
\label{fig:gamma}
\includegraphics[width=0.3\textwidth]{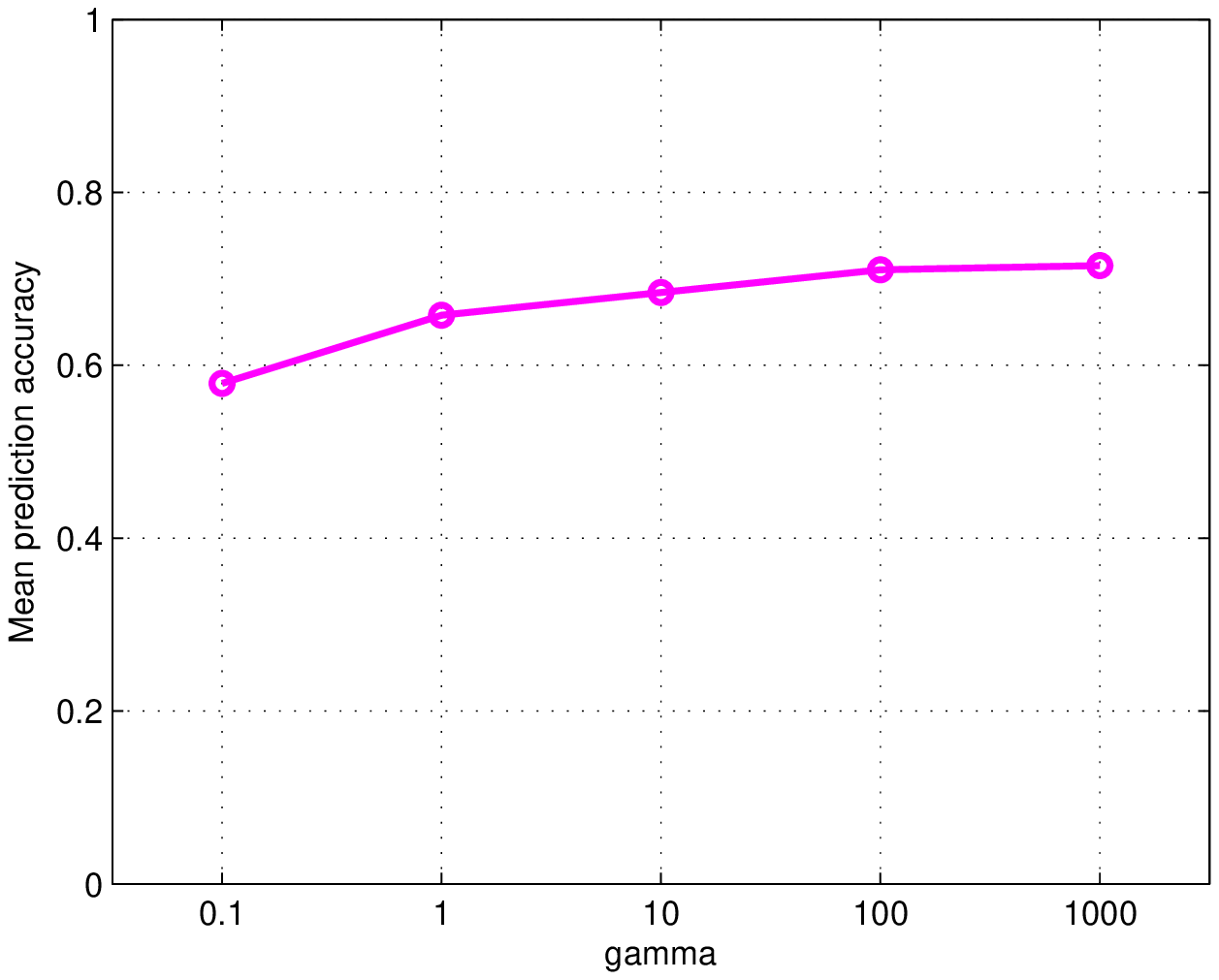}}
\\
\caption{Parameter sensitivity curves.}
  \label{fig:para}
\end{figure}

In the proposed formulation, there are three tradeoff parameters, $\alpha$, $\beta$, and $\gamma$. We plot the curve of mean prediction accuracies against different values of parameters, and show them in figure \ref{fig:para}. From figure \ref{fig:alpha} and \ref{fig:beta}, we can see the accuracy is stable to the parameter $\alpha$ and $\beta$.
From figure \ref{fig:gamma}, we can see a larger $\gamma$ leads to better classification performances.

\section{Conclusion and future works}
\label{sec:conclusion}

In this paper, we study the problem of using context to represent and classify data points. We propose to use a sparse linear combination of the data points in the context of a data point to represent itself. Moreover, to increase the discriminative ability of the new representation, we develop an supervised method to learn the sparse context by learning it and a classifier together in an unified optimization framework. Experiments on three benchmark data sets show its advantage over state-of-the-art context-based data representation and classification methods. In the future, we will extend the proposed method to applications of information security \cite{zhan2013characterizing,xu2013cross,xu2012stochastic,xu2014adaptive,xu2014evasion,xu2010trustworthy,zhan2014characterization},
bioinformatics \cite{wang2012mathematical,wang2010conceptual,wang2014computational,peng2015modeling,wang2012multiple,wang2012proclusensem,hu2009improving,zhang2010bioinformatics,hu2009improving}, computer vision \cite{wang2013joint,wang2012adaptive}, and big data analysis using high performance computing \cite{zhou2013exploring,wang2013using,li2013distributed,zhang2014lucas,gao2014sparse,zhang2013fpga,zhang2011frequent,zhang2013accelerating,zhang2011gpapriori,zhang2014lucas,li2013zht,zhao2014fusionfs,wang2014optimizing,wang2014next,zhou2013exploring,wangovercoming,wang2015towards}.


\begin{thebibliography}{10}
\providecommand{\url}[1]{{#1}}
\providecommand{\urlprefix}{URL }
\expandafter\ifx\csname urlstyle\endcsname\relax
  \providecommand{\doi}[1]{DOI~\discretionary{}{}{}#1}\else
  \providecommand{\doi}{DOI~\discretionary{}{}{}\begingroup
  \urlstyle{rm}\Url}\fi

\bibitem{Deng2009}
Deng, Q., Cai, A.: Svm-based loss differentiation mechanism in mobile ad hoc
  networks.
\newblock In: 2009 Global Mobile Congress, GMC 2009 (2009).
\newblock \doi{10.1109/GMC.2009.5295834}

\bibitem{gao2013laplacian}
Gao, S., Tsang, I.H., Chia, L.T.: Laplacian sparse coding, hypergraph laplacian
  sparse coding, and applications.
\newblock IEEE Transactions on Pattern Analysis and Machine Intelligence
  \textbf{35}(1), 92--104 (2013)

\bibitem{gao2010local}
Gao, S., Tsang, I.W., Chia, L.T., Zhao, P.: Local features are not
  lonely--laplacian sparse coding for image classification.
\newblock In: Computer Vision and Pattern Recognition (CVPR), 2010 IEEE
  Conference on, pp. 3555--3561. IEEE (2010)

\bibitem{gao2014sparse}
Gao, Y., Zhang, F., Bakos, J.D.: Sparse matrix-vector multiply on the keystone
  ii digital signal processor.
\newblock In: High Performance Extreme Computing Conference (HPEC), 2014 IEEE,
  pp. 1--6 (2014)

\bibitem{Guo20121893}
Guo, Z., Li, Q., You, J., Zhang, D., Liu, W.: Local directional derivative
  pattern for rotation invariant texture classification.
\newblock Neural Computing and Applications \textbf{21}(8), 1893--1904 (2012)

\bibitem{He2013793}
He, Y., Sang, N.: Multi-ring local binary patterns for rotation invariant
  texture classification.
\newblock Neural Computing and Applications \textbf{22}(3-4), 793--802 (2013)

\bibitem{hu2009improving}
Hu, J., Zhang, F.: Improving protein localization prediction using amino acid
  group based physichemical encoding.
\newblock In: Bioinformatics and Computational Biology, pp. 248--258 (2009)

\bibitem{Huang2014488}
Huang, F., Li, C., Lin, L.: Identifying gender of microblog users based on
  message mining.
\newblock Lecture Notes in Computer Science (including subseries Lecture Notes
  in Artificial Intelligence and Lecture Notes in Bioinformatics) \textbf{8485
  LNCS}, 488--493 (2014)

\bibitem{li2013distributed}
Li, T., Zhou, X., Brandstatter, K., Raicu, I.: Distributed key-value store on
  hpc and cloud systems.
\newblock In: 2nd Greater Chicago Area System Research Workshop (GCASR).
  Citeseer (2013)

\bibitem{li2013zht}
Li, T., Zhou, X., Brandstatter, K., Zhao, D., Wang, K., Rajendran, A., Zhang,
  Z., Raicu, I.: Zht: A light-weight reliable persistent dynamic scalable
  zero-hop distributed hash table.
\newblock In: Parallel \& Distributed Processing (IPDPS), 2013 IEEE 27th
  International Symposium on, pp. 775--787 (2013)

\bibitem{melacci2011laplacian}
Melacci, S., Belkin, M.: Laplacian support vector machines trained in the
  primal.
\newblock The Journal of Machine Learning Research \textbf{12}, 1149--1184
  (2011)

\bibitem{peng2015modeling}
Peng, B., Liu, Y., Zhou, Y., Yang, L., Zhang, G., Liu, Y.: Modeling
  nanoparticle targeting to a vascular surface in shear flow through diffusive
  particle dynamics.
\newblock Nanoscale Research Letters \textbf{10}(1), 235 (2015)

\bibitem{Tian20141007}
Tian, Y., Zhang, Q., Liu, D.: v-nonparallel support vector machine for pattern
  classification.
\newblock Neural Computing and Applications \textbf{25}(5), 1007--1020 (2014)

\bibitem{wang2012proclusensem}
Wang, J., Li, Y., Wang, Q., You, X., Man, J., Wang, C., Gao, X.: Proclusensem:
  predicting membrane protein types by fusing different modes of pseudo amino
  acid composition.
\newblock Computers in biology and medicine \textbf{42}(5), 564--574 (2012)

\bibitem{wang2012multiple}
Wang, J.J.Y., Bensmail, H., Gao, X.: Multiple graph regularized protein domain
  ranking.
\newblock BMC bioinformatics \textbf{13}(1), 307 (2012)

\bibitem{wang2013joint}
Wang, J.J.Y., Bensmail, H., Gao, X.: Joint learning and weighting of visual
  vocabulary for bag-of-feature based tissue classification.
\newblock Pattern Recognition \textbf{46}(12), 3249--3255 (2013)

\bibitem{wang2012adaptive}
Wang, J.Y., Almasri, I., Gao, X.: Adaptive graph regularized nonnegative matrix
  factorization via feature selection.
\newblock In: Pattern Recognition (ICPR), 2012 21st International Conference
  on, pp. 963--966 (2012)

\bibitem{wang2013using}
Wang, K., Kulkarni, A., Zhou, X., Lang, M., Raicu, I.: Using simulation to
  explore distributed key-value stores for exascale system services.
\newblock In: 2nd Greater Chicago Area System Research Workshop (GCASR) (2013)

\bibitem{wangovercoming}
Wang, K., Liu, N., Sadooghi, I., Yang, X., Zhou, X., Lang, M., Sun, X.H.,
  Raicu, I.: Overcoming hadoop scaling limitations through distributed task
  execution.
\newblock In: Proc. of the IEEE International Conference on Cluster Computing
  2015 (Cluster¡¯15) (2015)

\bibitem{wang2014next}
Wang, K., Zhou, X., Chen, H., Lang, M., Raicu, I.: Next generation job
  management systems for extreme-scale ensemble computing.
\newblock In: Proceedings of the 23rd international symposium on
  High-performance parallel and distributed computing, pp. 111--114 (2014)

\bibitem{wang2014optimizing}
Wang, K., Zhou, X., Li, T., Zhao, D., Lang, M., Raicu, I.: Optimizing load
  balancing and data-locality with data-aware scheduling.
\newblock In: Big Data (Big Data), 2014 IEEE International Conference on, pp.
  119--128 (2014)

\bibitem{wang2015towards}
Wang, K., Zhou, X., Qiao, K., Lang, M., McClelland, B., Raicu, I.: Towards
  scalable distributed workload manager with monitoring-based weakly consistent
  resource stealing.
\newblock In: Proceedings of the 24rd international symposium on
  High-performance parallel and distributed computing, pp. 219--222. ACM (2015)

\bibitem{wang2014computational}
Wang, S., Zhou, Y., Tan, J., Xu, J., Yang, J., Liu, Y.: Computational modeling
  of magnetic nanoparticle targeting to stent surface under high gradient
  field.
\newblock Computational mechanics \textbf{53}(3), 403--412 (2014)

\bibitem{wang2010conceptual}
Wang, Y., Han, H.C., Yang, J.Y., Lindsey, M.L., Jin, Y.: A conceptual cellular
  interaction model of left ventricular remodelling post-mi: dynamic network
  with exit-entry competition strategy.
\newblock BMC systems biology \textbf{4}(Suppl 1), S5 (2010)

\bibitem{wang2012mathematical}
Wang, Y., Yang, T., Ma, Y., Halade, G.V., Zhang, J., Lindsey, M.L., Jin, Y.F.:
  Mathematical modeling and stability analysis of macrophage activation in left
  ventricular remodeling post-myocardial infarction.
\newblock BMC genomics \textbf{13}(Suppl 6), S21 (2012)

\bibitem{wright2009robust}
Wright, J., Yang, A.Y., Ganesh, A., Sastry, S.S., Ma, Y.: Robust face
  recognition via sparse representation.
\newblock Pattern Analysis and Machine Intelligence, IEEE Transactions on
  \textbf{31}(2), 210--227 (2009)

\bibitem{xu2013cross}
Xu, L., Zhan, Z., Xu, S., Ye, K.: Cross-layer detection of malicious websites.
\newblock In: Proceedings of the third ACM conference on Data and application
  security and privacy, pp. 141--152. ACM (2013)

\bibitem{xu2014evasion}
Xu, L., Zhan, Z., Xu, S., Ye, K.: An evasion and counter-evasion study in
  malicious websites detection.
\newblock In: Communications and Network Security (CNS), 2014 IEEE Conference
  on, pp. 265--273. IEEE (2014)

\bibitem{xu2014adaptive}
Xu, S., Lu, W., Xu, L., Zhan, Z.: Adaptive epidemic dynamics in networks:
  Thresholds and control.
\newblock ACM Transactions on Autonomous and Adaptive Systems (TAAS)
  \textbf{8}(4), 19 (2014)

\bibitem{xu2012stochastic}
Xu, S., Lu, W., Zhan, Z.: A stochastic model of multivirus dynamics.
\newblock Dependable and Secure Computing, IEEE Transactions on \textbf{9}(1),
  30--45 (2012)

\bibitem{xu2010trustworthy}
Xu, S., Qian, H., Wang, F., Zhan, Z., Bertino, E., Sandhu, R.: Trustworthy
  information: concepts and mechanisms.
\newblock In: Web-Age Information Management, pp. 398--404. Springer (2010)

\bibitem{Xu20121205}
Xu, Y., Shen, F., Zhao, J.: An incremental learning vector quantization
  algorithm for pattern classification.
\newblock Neural Computing and Applications \textbf{21}(6), 1205--1215 (2012)

\bibitem{zhan2013characterizing}
Zhan, Z., Xu, M., Xu, S.: Characterizing honeypot-captured cyber attacks:
  Statistical framework and case study.
\newblock Information Forensics and Security, IEEE Transactions on
  \textbf{8}(11), 1775--1789 (2013)

\bibitem{zhan2014characterization}
Zhan, Z., Xu, M., Xu, S.: A characterization of cybersecurity posture from
  network telescope data.
\newblock In: Proceedings of the 6th international conference on trustworthy
  systems, Intrust, vol.~14 (2014)

\bibitem{zhang2014lucas}
Zhang, F., Gao, Y., Bakos, J.D.: Lucas-kanade optical flow estimation on the ti
  c66x digital signal processor.
\newblock In: High Performance Extreme Computing Conference (HPEC), 2014 IEEE,
  pp. 1--6 (2014)

\bibitem{zhang2009bayesian}
Zhang, F., Hu, J.: Bayesian classifier for anchored protein sorting discovery.
\newblock In: Bioinformatics and Biomedicine, 2009. BIBM'09. IEEE International
  Conference on, pp. 424--428 (2009)

\bibitem{zhang2010bioinformatics}
Zhang, F., Hu, J.: Bioinformatics analysis of physicochemical properties of
  protein sorting signals  (2010)

\bibitem{zhang2011gpapriori}
Zhang, F., Zhang, Y., Bakos, J.: Gpapriori: Gpu-accelerated frequent itemset
  mining.
\newblock In: Cluster Computing (CLUSTER), 2011 IEEE International Conference
  on, pp. 590--594 (2011)

\bibitem{zhang2013accelerating}
Zhang, F., Zhang, Y., Bakos, J.D.: Accelerating frequent itemset mining on
  graphics processing units.
\newblock The Journal of Supercomputing \textbf{66}(1), 94--117 (2013)

\bibitem{zhang2011frequent}
Zhang, Y., Zhang, F., Bakos, J.: Frequent itemset mining on large-scale shared
  memory machines.
\newblock In: Cluster Computing (CLUSTER), 2011 IEEE International Conference
  on, pp. 585--589 (2011)

\bibitem{zhang2013fpga}
Zhang, Y., Zhang, F., Jin, Z., Bakos, J.D.: An fpga-based accelerator for
  frequent itemset mining.
\newblock ACM Transactions on Reconfigurable Technology and Systems (TRETS)
  \textbf{6}(1), 2 (2013)

\bibitem{zhao2014fusionfs}
Zhao, D., Zhang, Z., Zhou, X., Li, T., Wang, K., Kimpe, D., Carns, P., Ross,
  R., Raicu, I.: Fusionfs: Toward supporting data-intensive scientific
  applications on extreme-scale high-performance computing systems.
\newblock In: Big Data (Big Data), 2014 IEEE International Conference on, pp.
  61--70 (2014)

\bibitem{zhou2013exploring}
Zhou, X., Chen, H., Wang, K., Lang, M., Raicu, I.: Exploring distributed
  resource allocation techniques in the slurm job management system.
\newblock Illinois Institute of Technology, Department of Computer Science,
  Technical Report  (2013)

\end{thebibliography}

\end{document}